\documentclass[10pt,twocolumn,letterpaper]{article}

\usepackage[pagenumbers]{cvpr} 

\usepackage{graphicx}
\usepackage{amsmath}
\usepackage{amssymb}
\usepackage{booktabs}
\usepackage{subcaption}
\usepackage{multirow}
\usepackage[symbol]{footmisc}
\usepackage[accsupp]{axessibility}

\newdimen\figrasterwd
\figrasterwd\textwidth

\usepackage[pagebackref,breaklinks,colorlinks]{hyperref}

\usepackage[capitalize]{cleveref}
\crefname{section}{Sec.}{Secs.}
\Crefname{section}{Section}{Sections}
\Crefname{table}{Table}{Tables}
\crefname{table}{Tab.}{Tabs.}

\newcommand{\minisection}[1]{\vspace{2mm}\noindent{\textbf{#1}}}

\def\cvprPaperID{10311} 
\def\confName{CVPR}
\def\confYear{2023}

\begin{document}

\title{Efficient Scale-Invariant Generator with Column-Row Entangled Pixel Synthesis}

\author{
Thuan Hoang Nguyen$^*$ \quad Thanh Van Le$^*$ \quad Anh Tran \\
VinAI Research, Hanoi, Vietnam \\
{\tt\small \{v.thuannh5,v.thanhlv19,v.anhtt152\}@vinai.io}
}

\maketitle

\footnotetext[1]{Equal contribution.}

\begin{abstract}
    Any-scale image synthesis offers an efficient and scalable solution to synthesize photo-realistic images at any scale, even going beyond 2K resolution. However, existing GAN-based solutions depend excessively on convolutions and a hierarchical architecture, which introduce inconsistency and the ``texture sticking'' issue when scaling the output resolution. From another perspective, INR-based generators are scale-equivariant by design, but their huge memory footprint and slow inference hinder these networks from being adopted in large-scale or real-time systems. In this work, we propose \textbf{C}olumn-\textbf{R}ow \textbf{E}ntangled \textbf{P}ixel \textbf{S}ynthesis (\textbf{CREPS}), a new generative model that is both efficient and scale-equivariant without using any spatial convolutions or coarse-to-fine design. To save memory footprint and make the system scalable, we employ a novel bi-line representation that decomposes layer-wise feature maps into separate ``thick'' column and row encodings. Experiments on various datasets, including FFHQ, LSUN-Church, MetFaces, and Flickr-Scenery, confirm CREPS' ability to synthesize scale-consistent and alias-free images at any arbitrary resolution with proper training and inference speed. Code is available at \url{https://github.com/VinAIResearch/CREPS}.
\end{abstract}


\section{Introduction}\label{sec:intro}

Generative Adversarial Networks (GANs) \cite{Ian14} are one of the most widely used structures for image generation and manipulation \cite{Weihao21, Amit22}. Previously, a GAN model could only generate images with a fixed scale and layout as defined in the training dataset. However, natural images come with varying resolutions and contain unstructured objects at diverse poses. Therefore, designing a generative model that can handle more flexible geometric configurations is gaining more attention in the machine-learning community. StyleGAN3 \cite{Tero21} already supports out-of-the-box translation and rotation with consistent and artifact-free outputs. Any-scale synthesis, however, remains under-explored.

\begin{figure}[t]
    \small
    \begin{center}
        \includegraphics[width=.477\textwidth]{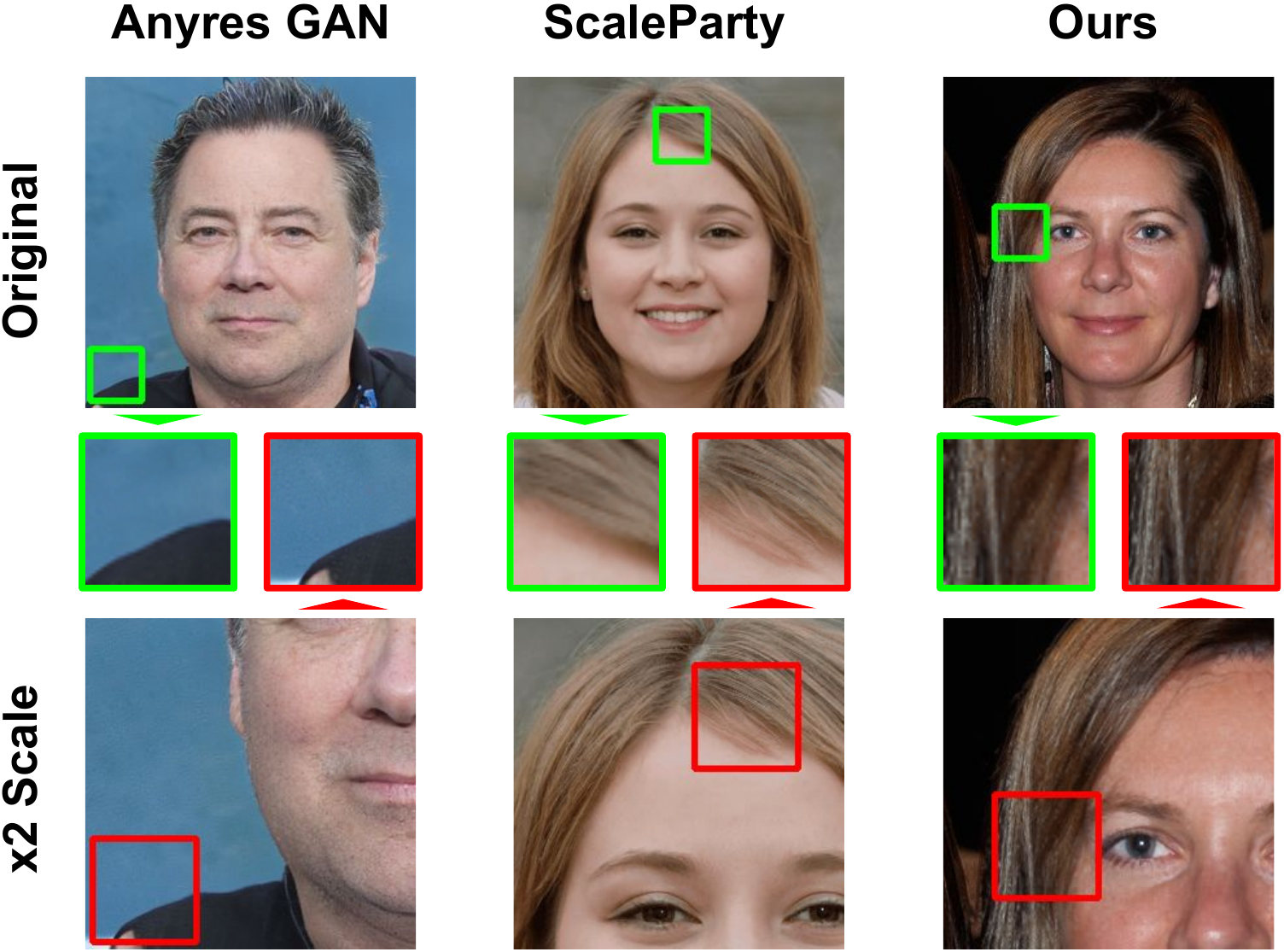}
    \end{center}
    \vskip -0.22in
    \caption{Previous any-scale image synthesis networks, including AnyresGAN \cite{chai2022anyresolution} and ScaleParty \cite{Evangelos22}, produce inconsistent image details when changing the output scale (see zoomed-in patches). In contrast, our proposed network can produce the same details but sharper when increasing the scale. Check the supplemental video mentioned in \cref{sec:scaling} for a clearer comparison.}
    \label{fig:teaser}
    \vskip -0.15in
\end{figure}

In this paper, we are interested in the task of arbitrary-scale image synthesis where a \textbf{single} generator can effortlessly synthesize images at many different scales while \textbf{strongly} preserving detail consistency. Such a model can be a promising research direction and bring many benefits. It enables synthesizing a high-resolution image from a lower-resolution training dataset. Hence, it eliminates the need for collecting and training models on high-resolution images, which is costly in storage, time, and computation resources. The output resolution can be ultra-high, e.g., $2048\times2048$, which is impossible for standard GAN models due to the limit of GPU memory. Any-scale image synthesis also allows geometric interactions like zooming in and out. Despite promising results, previous works on this topic, such as AnyresGAN \cite{chai2022anyresolution} and ScaleParty \cite{Evangelos22}, show strong inconsistency when scaling the output resolution (see \cref{fig:teaser}).

We investigate the GAN structures to find the potential cause of the inconsistency at image scaling. Traditional GAN models are based on convolutional generators \cite{Alec15, Andrew18, Tero20}, which introduce an implicit spatial bias that helps the model to produce high-quality images. Recently, Xu et al. \cite{Xu20} and Karras et al. \cite{Tero21} discovered that these positional priors can hamper the model's consistency when applying translations, rotations, or scalings. In order to combat this issue, many works introduce non-trivial changes such as sophisticated architecture re-design \cite{Tero21} or opt for a better training strategy and input positional encoding \cite{Evangelos22, chai2022anyresolution}. However, these are only partial remedies as the output pixels still depend on their surroundings, making it impossible for these models to produce consistent attributes of an object regardless of positions and scales.

In contrast to the traditional GANs, some recent methods are based on Implicit Neural Representation (INR) \cite{Ivan20, Anokhin20}. By predicting the color of each pixel separately, INR-based GANs can, in theory, synthesize objects in a spatial-arbitrary manner and still achieve comparable quality at small to medium resolution compared to convolution-based approach. However, these models' memory usage grows quadratically with the input resolution since all pixels have to be queried. Thus, there has been no existing work that can efficiently scale INR-GANs to resolutions higher than $1024$. To reduce training complexity, Anokhin et al. \cite{Anokhin20} employs a simple patch-based strategy where only a portion of pixels is generated and passed through the discriminator at a time. However, this approach unsurprisingly leads to poor results and inconsistency between patches.

Inspired by the latter approach, we aim to tackle the task of scale-consistent image generation with essential changes to StyleGAN2 \cite{Tero20}. Similar to Anokhin et al. \cite{Anokhin20}, we change the $3\times3$ convolutions to $1\times1$ ones and add a Fourier feature embedding \cite{Tancik20} at the input layer. Although these two changes alone already achieve our goal, it is still expensive to train in high-resolution settings. Thus, instead of using dense 2D features, our model relies on a novel thick bi-line representation, which largely reduces the training and inference complexity by using two low-rank features for row and column. Our network first regresses these row and column embeddings, then composes layer-wise intermediate 2D features, and finally fuses these maps to produce the final output. We name this novel structure \textbf{C}olumn-\textbf{R}ow \textbf{E}ntangled \textbf{P}ixel \textbf{S}ynthesis, or \textbf{CREPS} for short. 

We run a series of experiments on four datasets, including FFHQ, MetFaces, LSUN-Church, and Flickr-Scenery, to confirm the effectiveness of our proposed CREPS structure. Our model can synthesize images with quality comparable to the previous generative models like CIPS or StyleGAN2. While CIPS has trouble in training on images of resolutions more than $256\times256$, CREPS can sufficiently handle training data at resolutions $512\times512$ and $1024\times1024$. CREPS produces scale-equivariant images and keeps the object details unchanged when scaling the output resolution, unlike previous any-scale GANs such as AnyresGAN and ScaleParty. Using a CREPS model trained on $512\times512$ images, we still can generate near-realistic images at higher resolution. Finally, we demonstrate CREPS's ability to synthesize images with complex geometric transformations and distortions while preserving attribute consistency.

To summarize our contributions:
\begin{itemize}
    \item We propose a simple and elegant network equipped with only modulated linear layers and no upsampling layers in-between. It supports scale-consistent outputs for any-scale image synthesis. 
    \item To further improve efficiency, we introduce a thick bi-line representation, which decomposes 2D network features into two light-weight row and column embeddings. It significantly saves memory and computation costs compared with the full 2D-feature counterparts.
    \item We demonstrate competitive results for unconditional image synthesis on the FFHQ, LSUN-Church, MetFaces, and Flickr-Scenery datasets, along with the ability to generate each image at arbitrary scales with consistent details.
    \item Our CREPS models support complex geometric transformations and distortions. 
\end{itemize}

\section{Related Work}\label{sec:related}

\minisection{Generative Adversarial Networks.} Prior to denoising diffusion models \cite{Jonathan20, Jiamin20}, GANs \cite{Ian14} hold state-of-the-art results for image synthesis tasks. The popular GAN models can generate realistic images at a high resolution, commonly up to $1024\times1024$\cite{Andrew18, Tero19, Tero20, Tero20b, Tero21}. The promising results obtained by GANs have motivated several applications of computer graphics and visual content generation. However, these networks are only capable of generating images with same geometric configurations, e.g., center-located and face-forward objects. Recently, an exciting work StyleGAN3 \cite{Tero21} aimed to generalize GAN to arbitrary translation and rotation with consistent details, or Anycost GAN\cite{lin2021anycost} with multi-resolution generation. In the same vein, ScaleParty \cite{Evangelos22} and AnyresGAN \cite{chai2022anyresolution} extended StyleGAN2 and StyleGAN3 to support scaling and other geometric transformations by replacing learned input constant with suitable positional encoding and multi-scale training strategy. However, these works did not consider the scale consistency, and their images showed varied details as the output scale increases, illustrated in \cref{fig:teaser}.

\minisection{Implicit Neural Representation.} Typically, images are represented by a series of 2D arrays of values. However, it can be viewed as a continuous mapping from a 2D coordinate $(x, y) \in \mathbf{R}^2$ to the corresponding RGB value $(r, g, b) \in \mathbf{R}^3$ and the mapping can be parameterized as a black-box model. This coordinate-wise modeling has been used in a wide range of neural rendering tasks \cite{Ayush21, Vincent19, Ben20, Eric21}, where neural networks are used to provide an efficient and continuous representation of data compared with traditional methods. In the literature, implicit neural networks mainly utilize fully-connected layers as their building blocks. Unlike convolution or self-attention, such layers' receptive field size is exactly one; in other words, the output at every coordinate is independent of each other.

\minisection{INR-based GANs.} As the number of research increased, INR started to be used for generative tasks. These models soon inherited the success of GANs by employing the adversarial training manner. Generative radiance fields \cite{Schwarz20, Chan2022, Gu22, orel2022styleSDF} attempt to learn a view-consistent representation of 3D objects using implicit GAN. Despite all the success of INRs in 3D GANs, limited attention has been paid to utilizing the equivariance capability of fully-connected layers in 2D counterparts. The closest to our work are INR-GAN \cite{Ivan20} and CIPS \cite{Anokhin20}. Both these works use a grid of the target pixel coordinates as input for batch processing instead of passing each point individually. INR-GAN employs a multi-scale structure, which we will discuss later as a cause of scale inconsistency, while its uniform-scale versions have poor generation outputs. Meanwhile, CIPS does not need the multi-scale design thanks to its efficient weight modulation and expressive input embedding. The uniform-scale INR-GAN and CIPS disregard spatial convolutions in the generator and synthesize each pixel independently. However, their main goal is to investigate an alternative architecture that can compete with fully-convolutional GANs rather than paying attention to the equivariance characteristic of such models. They also struggle with expensive computation costs and memory usage using full-resolution 2D feature maps in processing.

\section{Proposed method}\label{sec:method}
This section describes our proposed CREPS structure. First, we recall the concept of any-scale image synthesis (\cref{sec:anyres}). Then, we revise two existing GAN structures that support scale-equivariant image synthesis (\cref{sec:no_spatial}). Next, we discuss how to reduce computation cost via the novel thick bi-line representation (\cref{sec:biline}). Finally, we describe the layer-wise feature composition scheme for improving the synthesis quality (\cref{sec:layerwise}).

\subsection{Any-scale image synthesis}\label{sec:anyres}
In this section, we introduce any-scale image synthesis as the task of generating images while enforcing consistency at different scales given a single model. One way we naturally come up with is generating an image at many scales altogether. MSG-GAN \cite{Karnewar19} is one of the earliest works in this approach. Instead of producing single output, MSG-GAN outputs an RGB image at each block of the generator, resulting in a mipmap representation \cite{Williams83}. However, this approach can only output pre-defined discrete scales, and there is no mechanism to guarantee scale consistency.

As such, we should consider injecting positional encoding $e$ as an additional input alongside the latent code into the generator. This approach is employed in some previous works \cite{Ivan20, Anokhin20, chai2022anyresolution, Evangelos22}, in which $e$ is a 2D grid of normalized $(x, y)$ coordinates. If $e$ is a regular grid, we can decompose it into two vectors for the row and column coordinates denoted as $e^r$ and $e^c$, respectively. The image generation process now becomes:
\begin{equation}
    I = G(z, e^r, e^c),
\end{equation}
with $G$ is the generative model and $z$ is the latent input. The decomposition from $e$ to $e^r$ and $e^c$ is more suitable to our thick bi-line representation, as later discussed. Doing so allows us to easily control the output's scale and other spatial properties via appropriate input encoding. However, naively adding positional input into an existing generator does not guarantee that the output image is equivariant to the change in the input coordinates. For example, when Karras et al. \cite{Tero21} replace the learnable constant in StyleGAN2's input layer with Fourier features (Config B), the ``texture sticking''' issue still occurs. Therefore, proper network design and training strategy should be examined to alleviate the output's geometric inconsistency. 

\begin{figure*}[t]
\centering
  \centering
  \parbox{\figrasterwd}{
    \parbox{.58\figrasterwd}{%
    \centering
      \subcaptionbox{Network structure\label{fig:network}}{\includegraphics[width=.77\hsize]{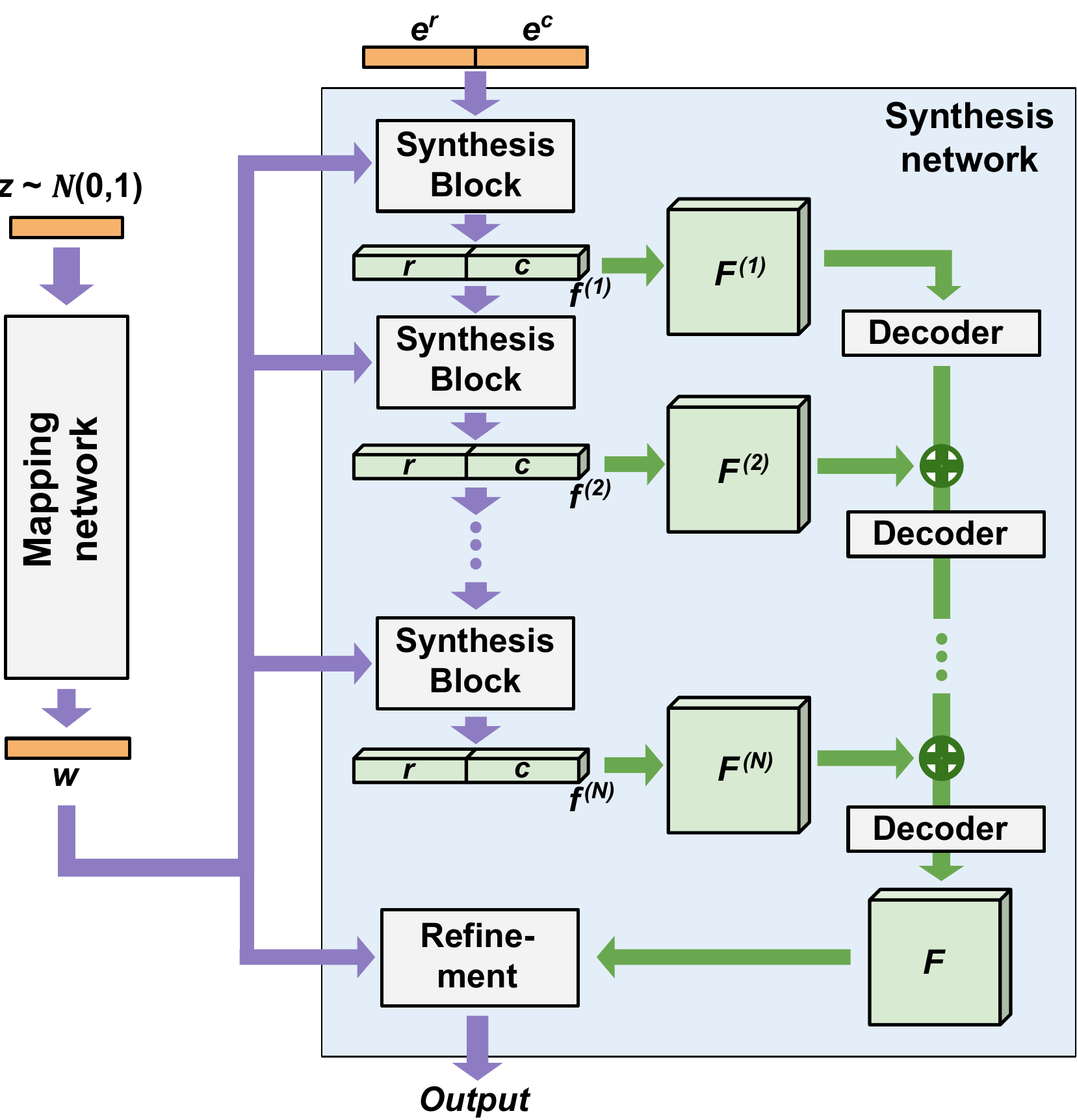}}
    }
    \hskip1em
    \parbox{.31\figrasterwd}{%
    \centering
      \subcaptionbox{Thick bi-line composition\label{fig:bilineConcept}}{\includegraphics[width=.7\hsize]{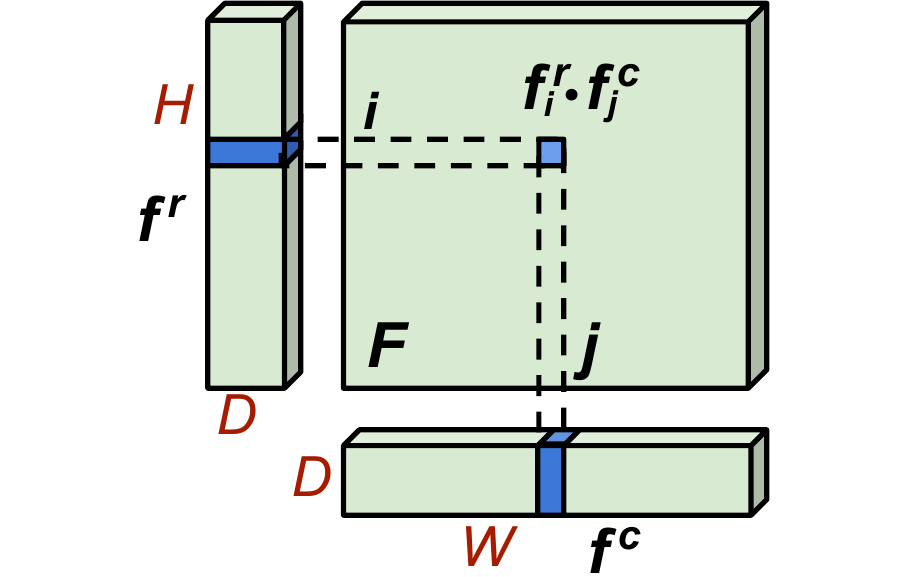}
      }
      \vskip1em
      \subcaptionbox{Refinement block\label{fig:refinement}}{\includegraphics[width=\hsize]{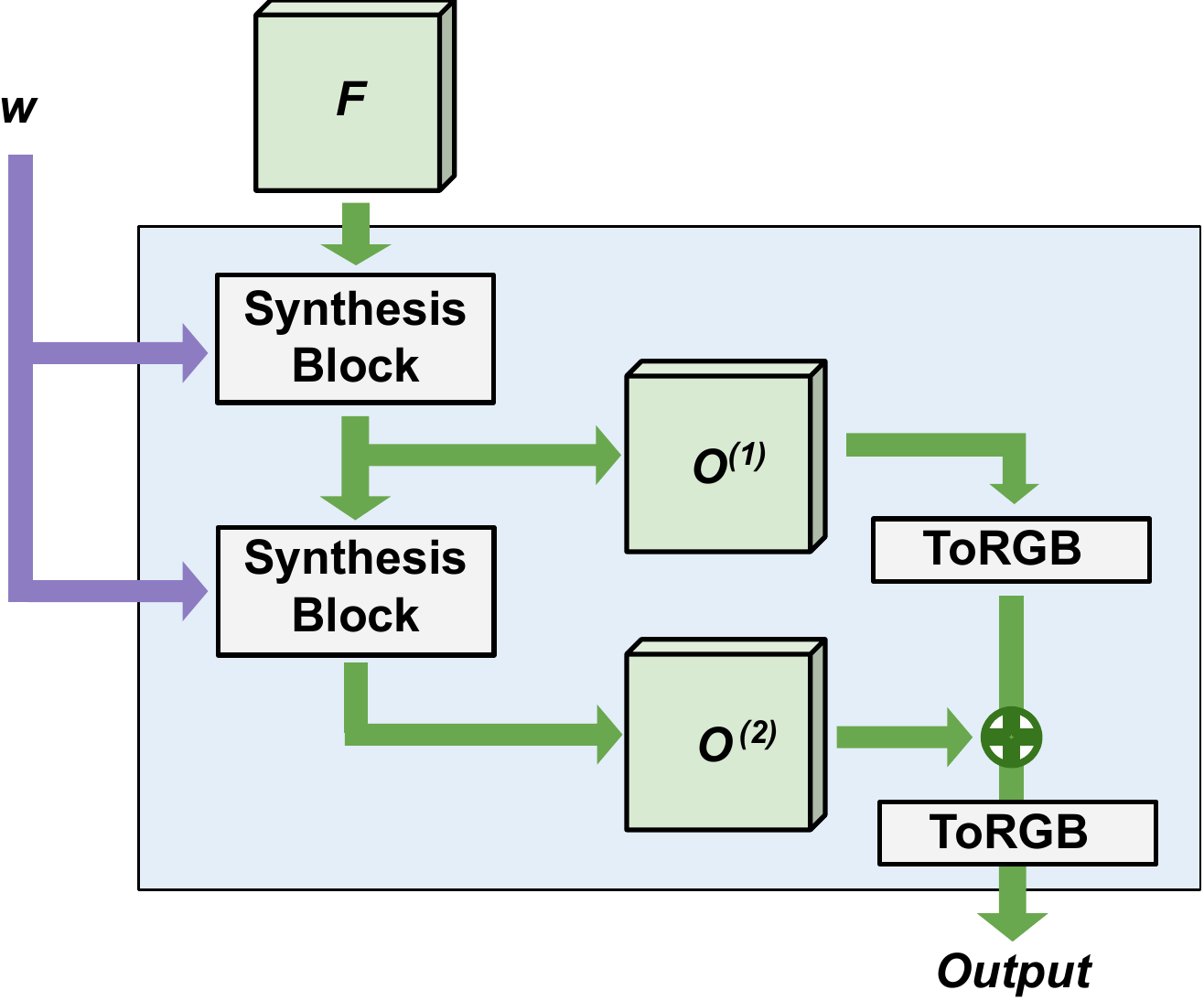}
      }  
    }
  }
\vspace{-3mm}
\caption{Our proposed CREPS structure}
\label{fig:overview}
\end{figure*}

\begin{table*}[t]
\centering
\addtolength{\tabcolsep}{+2pt}  
\begin{tabular}{cccccccc}
\toprule
\multirow{2.5}{*}{Resolution} & \multirow{2.5}{*}{Batch size} & \multicolumn{3}{c}{Memory Usage}              & \multicolumn{3}{c}{Running time}                       \\
\cmidrule(lr){3-5}\cmidrule(lr){6-8}
& & StyleGAN2 & CIPS & Ours  & StyleGAN2 & CIPS & Ours \\ 
\midrule
\multirow{2}{*}{$256\times256$} & 1 & 1.5GB & 3.3GB & 2.3GB & 0.04s & 0.06s & 0.03s \\
& 4 & 2.5GB & 10.2GB & 5.2GB & 0.05s & 0.23s & 0.06s \\
\midrule
\multirow{2}{*}{$512\times512$} & 1 & 1.7GB & 10.4GB & 4.5GB & 0.04s & 0.21s & 0.05s \\
& 4 & 3.4GB & OOM & 14.6GB & 0.06s & OOM & 0.16s \\
\bottomrule
\end{tabular}
\addtolength{\tabcolsep}{-2pt}  
\centering
\vspace{-2mm}
\caption{Memory usage and running time comparison between StyleGAN2, CIPS and our method. OOM means out-of-memory. \label{tab:cips_vs_stylegan2}}
\vspace{-4mm}
\end{table*}

\subsection{Removing coarse-to-fine design and spatial convolution}\label{sec:no_spatial}
We investigate two network structures that support any resolution image generation, including AnyresGAN \cite{chai2022anyresolution} and CIPS \cite{Anokhin20}, when keeping the same latent input but gradually increasing the output resolution. The former is built upon StyleGAN3 \cite{Tero21} with additional scale information concatenated with the latent code and a multi-scale training scheme. In contrast, the latter changes the 3x3 convolution of StyleGAN2 with a point-wise one and adds learnable Fourier features at the beginning. Both models are capable of multi-scale generation, but they have different behaviors that we will discuss below.

As illustrated in \cref{fig:teaser}, while having good photo-realism, AnyresGAN produces different image details at different scales. This can be explained by the fact that AnyresGAN, similar to most other GAN-based works, relies on spatial convolutions, such as 2D convolution with kernel size $3\times3$ and upsample layers. When changing the output resolution, the neighbor pixels at each location change, greatly varying the output of this spatial-convolution-based network. 

On the other hand, CIPS keeps the output image's details nearly same regardless of resolution, thanks to its spatial-free building operators. CIPS, however, is very computationally expensive; this can be clearly shown in \cref{tab:cips_vs_stylegan2}. When measured on a single NVIDIA V100 GPU (32 GB) and all models have comparable number of parameters, it runs slower than StyleGAN2 as well as requires much more memory or even gets an out-of-memory (OOM) error when running at $512 \times 512$ resolution. This makes CIPS inapplicable to use for learning fine details from high-resolution datasets. Moreover, it is worth noting that our method achieve the best trade-off between speed and memory.

Based on the above observations, we implement CREPS without any spatial convolutions or coarse-to-fine design. Starting with StyleGAN2 \cite{Tero20}, which consists of a mapping network and a generator, we remove all upsampling operators and replace all spatial convolutions with $1\times1$ convolutions, which are equivalent to pixel-wise fully-connected layers. Next, we replace the constant in the first synthesis block with Fourier encodings of the input coordinate row and column $e^r$ and $e^c$. This design is quite similar to CIPS, with only two minor differences. Firstly, the dense 2D grid input is now split into two vectors representing the row and column. Secondly, we do not combine learned input constant with the Fourier feature like CIPS did, making our model simpler and more memory-friendly. While this initial network guarantees any-scale image synthesis with consistent image details, it faces the same memory issue as CIPS. We will discuss next how to solve this issue effectively.


\subsection{Thick bi-line representation}\label{sec:biline}
Inspired by the tri-plane representation in \cite{Chan2022}, we propose to decompose each feature with 2D spatial dimensions into a column and a row embedding for a memory-efficient representation. For simplicity, let us drop the first two dimensions for the batch size $B$ and the number of channel $C$, which are the same and element-wise processed for both the feature map and the mentioned embeddings. Let us denote the feature map as $F \in \mathbb{R}^{H \times W}$, with $H$ and $W$ as the height and width, respectively. We can decompose $F$ to a row embedding $f^r$ and a column embedding $f^c$. In the simplest form, $f^r$ and $f^c$ are 1D vectors with the lengths $H$ and $W$, respectively. Each pixel in the feature map $F_{ij}$, with $i$ and $j$ as the row and the column indices, can be computed as the product of the corresponding elements in $f^r$ and $f^c$:
\begin{equation}
    F_{ij} = f^r_i f^c_j.
\end{equation}
We call this representation ``bi-line'', which significantly reduces the memory usage and computation cost and allows the network to learn with high-resolution data. However, we found that this simple representation had a limited capacity and could not define complex structures. To enrich its representation power, we ``thicken'' the embeddings by adding an extra, short dimension. The revised $f^r$ and $f^c$ now have the shapes $H \times D$ and $W \times D$, respectively, where $D \ll min(H, W)$ is a uniform embedding ``thickness''. The composing feature element $F_{ij}$ is now the dot product of the corresponding elements in $f^r$ and $f^c$:
\begin{equation}
    F_{ij} = f^r_i \cdot f^c_j = \sum_{d=1}^D f^r_{id} f^c_{jd}.
    \label{eq:thick_biline}
\end{equation}
This composition process is illustrated in \cref{fig:bilineConcept}. In another perspective, this can be considered as sum of $D$ different bi-line compositions. We call it ``thick bi-line'' representation. 

\begin{figure}[t]
    \small
    \begin{center}
        \includegraphics[width=.477\textwidth]{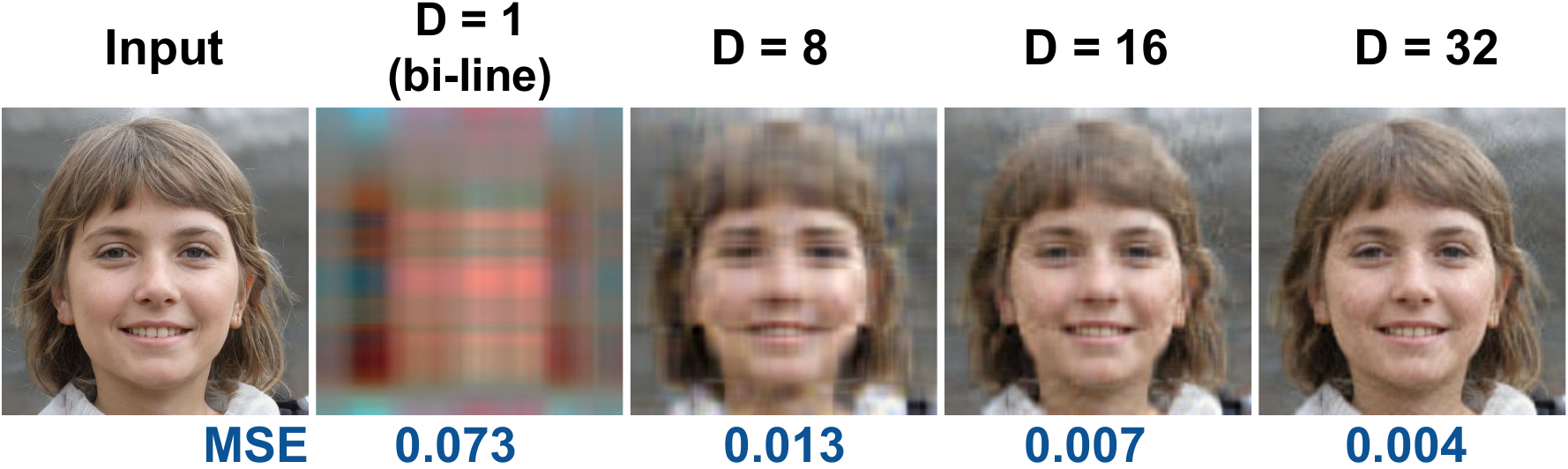}
    \end{center}
    \vskip -0.22in
    \caption{Fitting an input image to the thick bi-line representation in the image space.}
    \label{fig:biline}
    \vskip -0.15in
\end{figure}

In \cref{fig:biline}, we provide a toy example illustrating the capacity of the proposed thick bi-line representation. Given an input image at resolution $512\times512\times3$, we fit it into the proposed bi-line representation in the image space by optimizing a row and a column embedding of shape $512\times D\times3$. Note that each channel is optimized independently. As can be seen, with the naive bi-linear composition ($D=1$), the reconstructed image is just a simple, incomprehensible grid. By adding just a small thickness $D = 8$, we can capture the essential image content, recover the subject's identity, and reduce the MSE almost 6 times. When using $D=32$, we nearly recover the original image with only subtle pixel noise. Note that the row and column embeddings only take 1.56\% of the original image size when $D=8$ and 12.5\% when $D=32$. This experiment confirms the efficacy of our proposed thick bi-line decomposition. Also, while this representation does not capture all details of the complex input image, it is more sufficient when modeling the over-parameterized feature space.

\subsection{Layer-wise feature composition}\label{sec:layerwise}
In CREPS, we assume the target output is square, i.e., $H = W$. Hence, we can concatenate the row and column embeddings to a single tensor $f = [f^r, f^c] \in \mathbb{R}^{H \times 2D}$. Initially, we implement CREPS by revising StyleGAN2's code to predict $f$ from the latent input $w$ via $N$ synthesis blocks. The network then splits $f$ to get the row and column codes, perform the feature composition defined in \cref{eq:thick_biline} to get a feature map $F$. This feature map will be passed to a simple refinement module (\cref{fig:refinement}) with 2 synthesis blocks to produce the output image. For efficient memory and computation cost, we only employ a small thickness value $D = 8$.

We found this initial design needed to be more efficient to catch up with the generation quality of StyleGAN and CIPS. It performed feature composition once near the end of the image synthesis process; thus, the model power was bounded by the capacity of the thick bi-line representation. Instead, we revise our solution by employing a layer-wise feature composition scheme. Specially, at each layer with index $l \in [1..N]$, we extract the intermediate row and column embedding $f^{(l)}$. We can split $f^{(l)}$ and compose an intermediate feature map $F^{(l)}$, following \cref{eq:thick_biline}. Then, the intermediate maps across layers are fused to get the final map $F$. This scheme enriches the representation power, similar to when increasing $D$ while using less memory.

The fusion scheme is also important. Intuitively, we can set $F$ as the sum of the intermediate maps $\{F^{(l)}\}_{l=\overline{1,N}}$. However, this formulation treats the maps equally, and we find it undesirable. Let us call back the StyleGAN models' behavior. Thanks to the coarse-to-fine design, their early layers learn to capture the global shape, while the later layers learn to synthesize fine details. Since CREPS has no coarse-to-fine structure, it is hard to control which aspect of the output image each layer can learn. Hence, we propose adding asymmetry to the feature map fusion process: the feature maps at earlier layers are processed ``deeper'' than those at later layers. We hope it implicitly guides the layers to learn information from global to regional order, similar to StyleGAN. To do so, we introduce at each layer with index $l$ a narrow decoder,  denoted as $\pi^{(l)}$. The process to fuse the intermediate maps $\{F^{(l)}\}_{l=\overline{1,N}}$ is defined as following:
\begin{align}
    &E^{(1)} & = \; & F^{(1)} \label{eq:fuse_1},\\
    &E^{(l+1)} & = \; & \pi^{(l)}(E^{(l)}) + F^{(l+1)} \quad  \forall l \in [1, N-1]  \label{eq:fuse_2},\\
    &F & = \;  & \pi^N(E^N)  \label{eq:fuse_3},
\end{align}
with $E^{(l)}$ records the fused feature map at the $l^{th}$ layer. In our implementation, each decoder consists of pixel-wise fully-connected layers with Leaky-RELU activations. \cref{fig:network} illustrates our proposed network structure, while \cref{tab:cips_vs_stylegan2} illustrates the efficiency of our proposed structure in memory usage and running time.

\begin{figure}[t]
    \small
    \begin{center}
        \includegraphics[width=.47\textwidth]{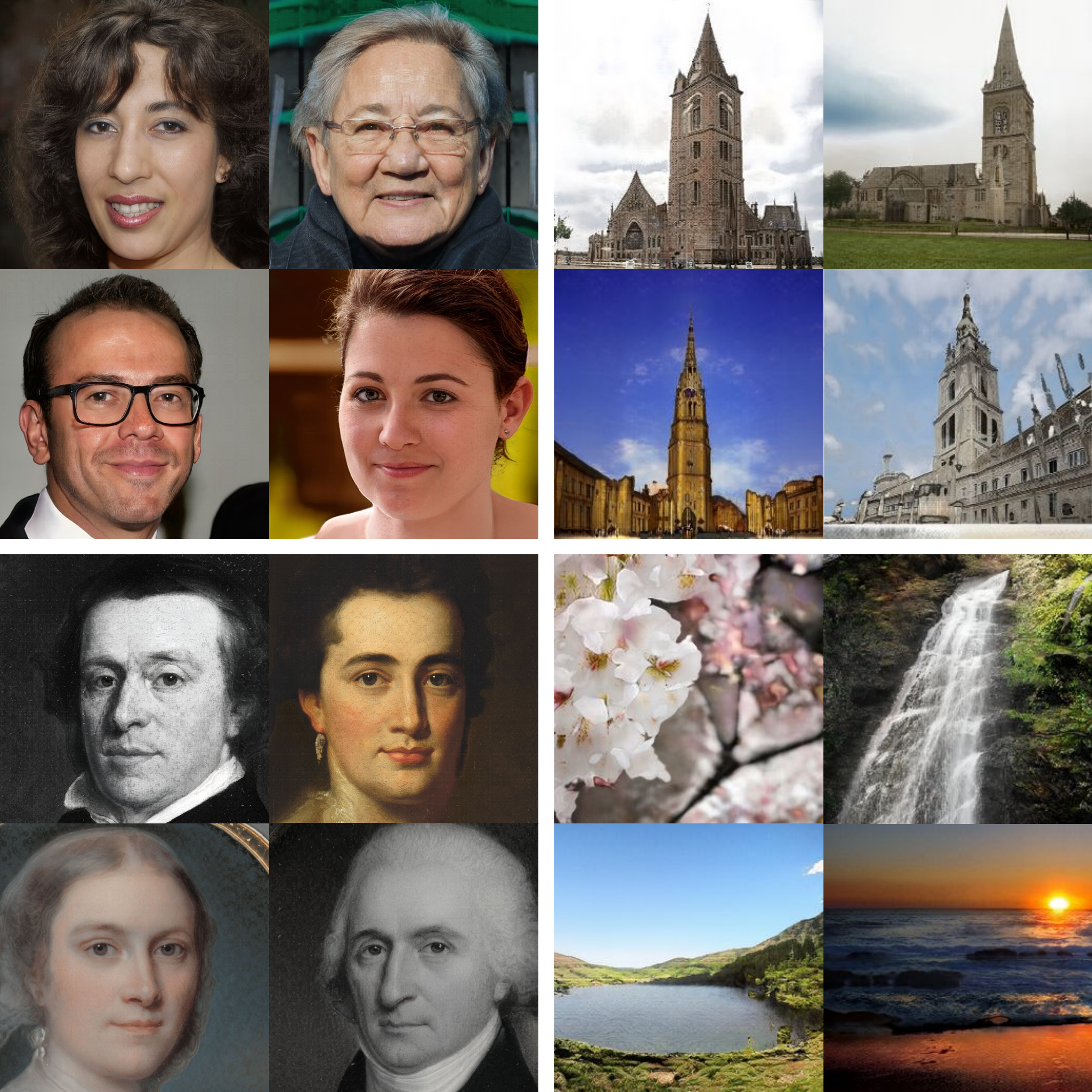}
    \end{center}
    \vskip -0.22in
    \caption{Sample images with our models trained on FFHQ (upper-left), LSUN-Church (upper-right), MetFace (bottom-left), and Flickr-Scenery (bottom-right).}
    \label{fig:qualitative}
    \vskip -0.15in
\end{figure}


\section{Experiments}\label{sec:exp}
\subsection{Experimental setup}\label{sec:exp_setup}

\minisection{Datasets.} We conduct experiments on the common datasets when benchmarking CREPS, including FFHQ, MetFaces, LSUN-Church, and Flickr-Scenery. FFHQ dataset contains 70k high-quality, diverse human faces collected from Flickr. We will use the FFHQ images with resolution $512\times512$. MetFaces is a small dataset of face drawings extracted from the collection of the Metropolitan Museum of Art,  with a total of 1336 images at resolution $1024\times1024$. LSUN-Church consists of 126k outdoor photographs of churches at the resolution $256\times256$. Finally, Flickr-Scenery \cite{InOut} is a landscape-centric dataset collected on Flickr with 50k images at resolution $256\times256$.


\minisection{Implementation.} We use StyleGAN2 network design as a reference to implement CREPS. Except for the refinement module, our generator consists of 6 (for the target resolution 256) to 8 synthesis blocks (for the resolution 1024) and the same number of decoder blocks. We replace all modulated convolution layers in StyleGAN2 with modulated fully-connected ones. Unlike StyleGAN2, the output of each block is not an RGB image but a 32-channel bi-line feature with thickness $D = 8$. Each decoder is a stack of $P = 4$ pixel-wise fully-connected layers, with the channel widths ranging from $32$ to $128$. This setting is applied for all experiments, except for our ablation study. Similar to StyleGAN3 and CIPS, we turn off style mixing regularization. Besides that, we kept most of the other components unchanged, including the mapping network, discriminator, path length regularization, and $R_1$ gradient penalty. 

\minisection{Training.} For FFHQ and LSUN-Church, our networks were trained from scratch until convergence. To verify the flexibility and scale consistency of CREPS on higher-resolution image synthesis, we increase the length of its coordinate input to generate images at resolution $1024\times1024$ on the FFHQ dataset. We also test the adaptability of our network on domain shift by applying transfer learning from the weights trained on FFHQ to MetFaces. Our networks were trained by Adam optimizer with learning rate $2\times10^{-3}$ and hyperparameters $\beta_{0} = 0$, $\beta_{1} = 0.99$, and $\epsilon = 10^{-8}$. We use 4 NVIDIA A100 40GB GPUs for training all models.

\begin{table*}[t]
\centering
\begin{tabular}{|c|c|c|c|c|c|}
\hline
Generator      & FFHQ-512    &     FFHQ-1024       & LSUN Church-256 &  MetFaces-1024 & Scenery-256 \\ \hline
StyleGANv2     & 3.41        &     2.84            &    3.86         &    18.22$^*$   &   6.40  \\ \hline
CIPS           & 6.18        &     10.07$^\dag$    &    2.92         &      OOM       &    8.49   \\ \hline
CREPS (ours)   & 4.43        &     4.09$^\ddag$    &    5.50         &    20.52       &    7.21    \\ \hline
\end{tabular}
\centering
\vspace{-2mm}
\caption{Comparison of our method against other works in FID metric. OOM means out-of-memory.   
`$^*$' means the result is taken from StyleGAN2-Ada paper \cite{Tero20b}. `$^\dag$' means the model is provided without releasing its progressive training code. `$^\ddag$' means the result is obtained by scaling the output resolution of the FFHQ-512 model. \label{tab:quantitative}}
\vspace{-3mm}
\end{table*}

\subsection{Image generation}\label{sec:img_gen}

\cref{tab:quantitative} compares the quality of images generated by our CREPS models with the standard spatial-convolution-based StyleGAN2 and the only scale-consistent any-scale image generation technique CIPS, using the Frechet Inception Distance (FID) score.

At resolution $512\times512$ on FFHQ, our model achieves the FID score of 4.43, which is much better than the score from CIPS (6.18) and not far from StyleGAN2 (3.41). We can also use this model to generate images at resolution $1024\times1024$ without retraining and achieve a better FID score (4.09). We found that CIPS cannot be trained for this resolution due to its expensive memory usage, even with training batch size 1, when using its official code. However, the authors provided a pretrained model for FFHQ-1024 using a progressive training scheme (no released code). This CIPS model has an FID score of 10.07, much worse than ours. This confirms the superiority of our method over its scale-consistent image generation counterpart.

On the MetFaces dataset, CREPS's FID score is 20.52, which is quite close to the score of StyleGAN2-Ada (18.22). As mentioned, CIPS fails to train on this $1024\times1024$ resolution using its official code. It confirms that bi-line representation does not constrain the adaptability of our model.

On LSUN-Church and Flickr-Scenery, although the unstructured and diverse images in these datasets are intuitively adverse to column and row decomposition, CREPS obtains good results with only a small gap compared with StyleGAN2's ones. Note that CIPS achieves a surprisingly good result on LSUN-Church; it surpasses not only CREPS but also StyleGAN2 in this setting.


\cref{fig:qualitative} provides some samples synthesized by our networks on the benchmark datasets. As can be seen, CREPS produces highly realistic images in all cases.

\subsection{Generate arbitrary-scale images}\label{sec:ultra}
While our models are trained on images with resolutions from $256\times256$ to $1024\times1024$, they can generate images at any scale. One way is that we simply scale the length of $e^r$ and $e^c$, and the output size is changed accordingly, thanks to our network design. With a V100 GPU (32GB), our models can generate an image up to resolution $3687\times3687$ in a single run. Or we can generate an image patch-by-patch with suitable coordinate inputs, then combine them together into a single gigantic image with no upper limit in the output size. We provide some images generated at 6K resolution at \href{https://drive.google.com/drive/folders/14tvr9x2JzMkCF1CWNbMtc0gDLuTigWv7?usp=share_link}{here}. While these images are not as sharp as real-world ultra-high-resolution images, they are much sharper than the ones generated at resolution $512\times512$ and then upscaled with Lanczos resampling.

\begin{table}[t]
\centering
\begin{tabular}{|c|c|c|c|}
\hline
           & PSNR $\uparrow$                & SSIM $\uparrow$              & LPIPS $\downarrow$                       \\ \hline
AnyresGAN  & 24.19          & 0.73          & 0.07          \\ \hline
ScaleParty & 24.50          & 0.70          & 0.08          \\ \hline
CIPS       & 33.33          & 0.93          & 0.05          \\ \hline
CREPS      & \textbf{34.65} & \textbf{0.96} & \textbf{0.01} \\ \hline
\end{tabular}
\centering
\vspace{-2mm}
\caption{Scale consistency comparison of our method against three other works on PSNR, SSIM and LPIP. The best scores are \textbf{bold}. \label{tab:scale_consistency}}
    \vskip -0.05in
\end{table}

\begin{figure}[t]
    \small
    \begin{center}
        \includegraphics[width=.47\textwidth]{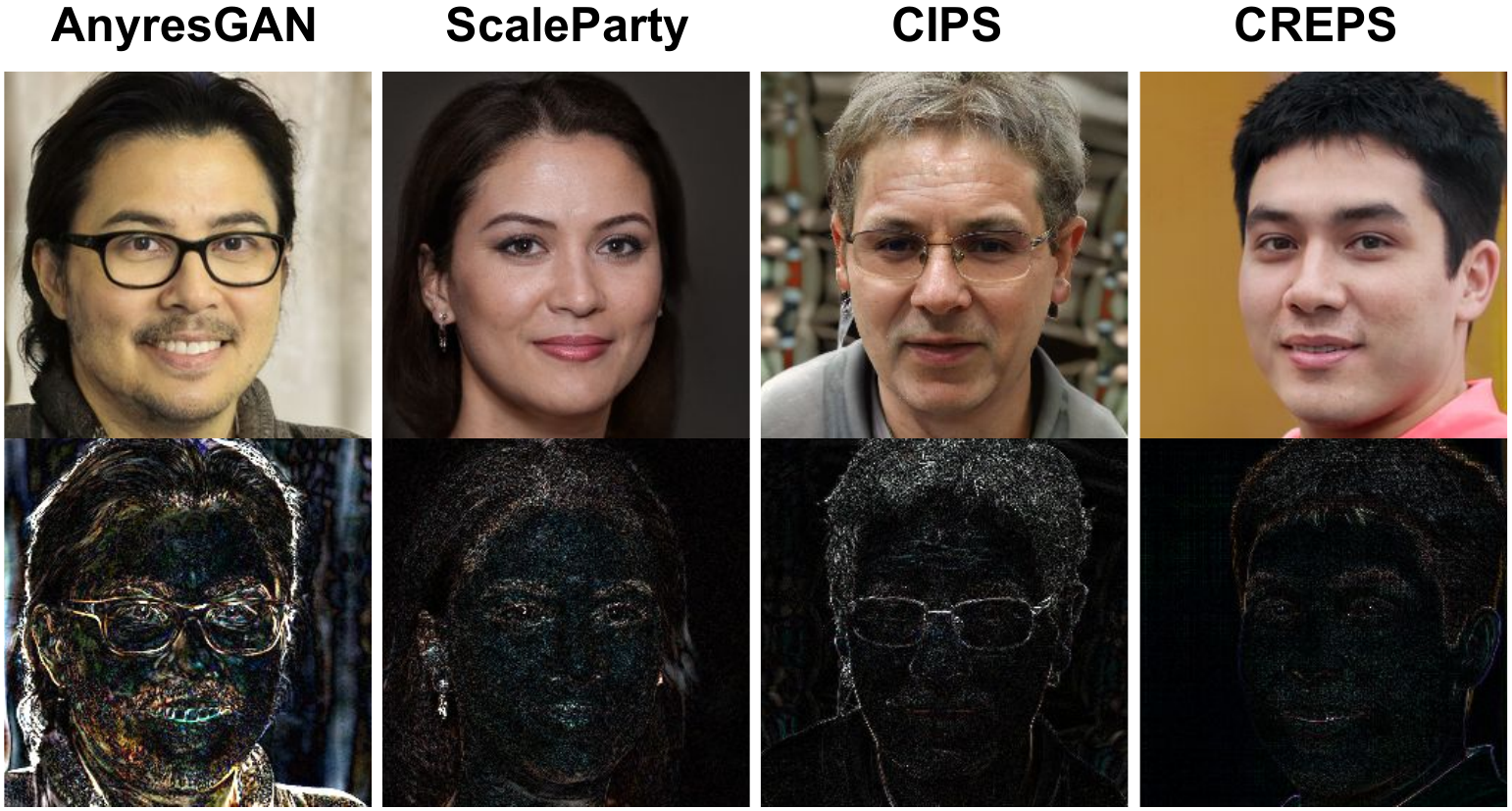}
    \end{center}
    \vskip -0.22in
    \caption{Qualitative results for the scale consistency experiment. For each method, we provide a sample generated $256\times256$ image (top) and the magnified ($\times10$) residual map between it and the $512\times512$ rescaled version (bottom).}
    \label{fig:scaleCon}
    \vskip -0.15in
\end{figure}

\subsection{Image scaling consistency}\label{sec:scaling}
In this section, we evaluate the scale consistency of images produced by CREPS and other methods, including AnyresGAN \cite{chai2022anyresolution}, ScaleParty \cite{Evangelos22}, and CIPS \cite{Anokhin20}. We run this experiment using models trained on the FFHQ dataset.
For each model, we first randomly generate 10k images at resolution $256 \times 256$ (first set). We then generate images with the same latent codes but at resolution $512 \times 512$ and downsample them to $256 \times 256$ (second set). The images in two sets are expected to be the same. Hence, we can compare two sets, with standard metrics such as PSNR, SSIM, and LPIPS, to measure each model's scale equivariance.

Note that for ScaleParty and AnyresGAN, the pretrained weights are already trained with different resolutions at once, so we directly use their provided version. CIPS, however, is trained in a single-scale setting, so we use the available weights trained at the highest resolution ($1024 \times 1024$) but pass the input coordinate with size $512 \times 512$ to synthesis image at resolution $512$. As for CREPS, we simply use the weight trained at resolution $512\times512$.

 We report the qualitative and quantitative results in \cref{fig:scaleCon} and \cref{tab:scale_consistency}. As can be seen, it is clear that CREPS achieves the best scale consistency, while convolution-based models like ScaleParty and AnyresGAN perform poorly.

Additionally, we provide a scale-consistency comparison video of CREPS with previous any-scale synthesis architectures, including AnyresGAN \cite{chai2022anyresolution}, ScaleParty \cite{Evangelos22}, and CIPS \cite{Anokhin20} at \href{https://drive.google.com/file/d/1PjMX90L_p74OBUkvADm57s7LMUh-iVgR/view}{here}. Note that, for a fair comparison, we use the provided codes from each method to produce the video except for ScaleParty, where we obtain the video directly from their codebase. For clearer visualization, we highlight the crop with the largest changes in AnyresGAN's output with a blue square.

\begin{table}[t]
\centering
\begin{tabular}{|l|c|cc|}
\hline
Configuration               & FID   & Memory & Time \\ \hline
CREP-NB                     & 5.98  & 2.7GB        & 0.13s        \\
+ bi-line and d=1           & 11.37 & 1.5GB        & 0.02s        \\
+ bi-line and d=8           & 8.23  & 1.6GB        & 0.03s        \\
+ no decoder and d=8        & 6.91  & 1.7GB        & 0.03s        \\
+ multiple decoders and d=4 & 6.46  & 1.6GB        & 0.03s        \\
+ multiple decoders and d=8 & 4.66  & 1.7GB        & 0.04s        \\ \hline
CIPS                        & 7.08  & 3.5GB        & 0.05s        \\ \hline
\end{tabular}
\centering
\vspace{-2mm}
\caption{Effects of the modifications of CREPS on the FFHQ dataset in terms of FID score, memory usage, and running time. \label{tab:ablate}}
\vspace{-5mm}
\end{table}

\subsection{Ablation studies}\label{sec:ablation}
To better understand our proposed techniques, we analyze the effect of different parts of CREPS on the FFHQ datasets. We first consider a no-bi-line version of CREPS as a baseline (referred to as \textit{CREPS-NB}), with the decoder layers removed and dense 2D used as input. We then apply bi-line decomposition but fuse the bi-line features only once at the end, with gradually increased thickness. Lastly, we add multiple decoders for layer-wise feature composition as introduced in \cref{sec:layerwise}. Because of limited time and computational resources, we only evaluate on $128 \times 128$ resolution and all of our models were trained for maximum of 2 days. 

As the results in \cref{tab:ablate} show, CIPS performs worst in all three aspects compared with most of our models. Simply adding the bi-line with a single decoder at the end nearly halves the memory costs, but the FID score is still behind CREPS-NB even when the thickness is increased to $d=8$. However, multiple decoders can help bring the image quality back to the level of CREPS-NB and even better with $d=8$. In all settings, it can be clearly seen that we easily boost the FID score when increasing the thickness. Moreover, we also omit the decoder $\pi$ between synthesis blocks and simplify the fusion scheme to $E^{(l+1)} = \; E^{(l)} + F^{(l+1)}$, which leads to even worse FID score than the smaller config with multiple decoders and d=4. These observations prove the importance and effectiveness of our proposed techniques. Remarkably, while the decoders seem compute-intensive, they are actually lightweight due to their narrow width compared with other layers, causing only small increases in memory and time.

\begin{figure}[t]
    \small
    \begin{center}
        \includegraphics[width=.477\textwidth]{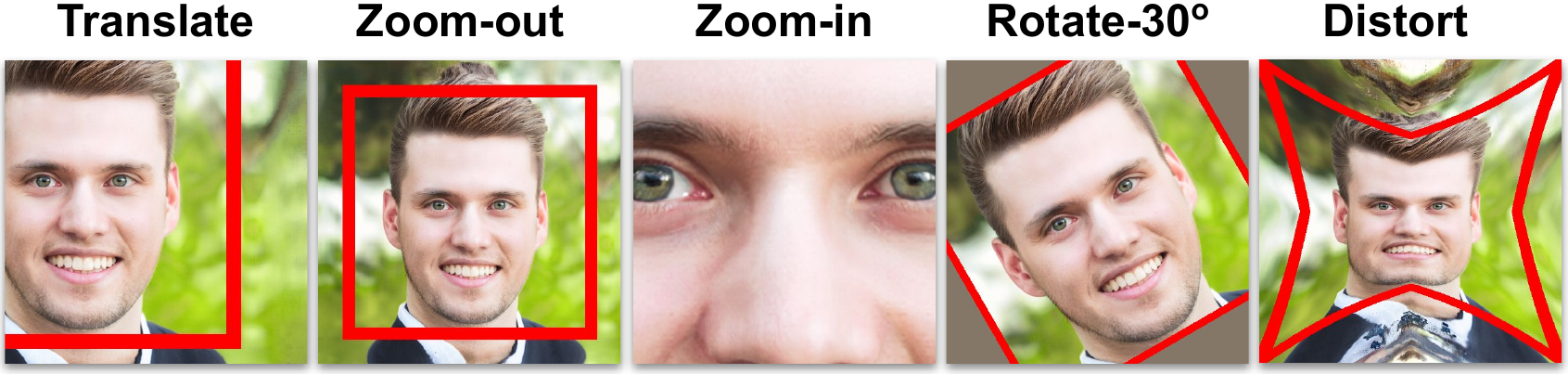}
    \end{center}
    \vskip -0.22in
    \caption{Geometric transformations on the same target image (FFHQ-512 model) by changing the input coordinates. We mark the original image boundary using a red rectangle.}
    \label{fig:geo_transform}
    \vspace{-2mm}
\end{figure}

\subsection{Simple and complex geometric transformation}\label{sec:distortion}
Our CREPS model can support various geometric transformations on the same target image by keeping the input latent code $z$ but changing the input coordinates $e^r$ and $e^c$. We can translate the image by adding coordinate shifts $\delta_y$ and $\delta_x$ to the row and column input coordinates, respectively. We can also multiply these input coordinates by the same constant $s > 1$ for zooming out or divide them by $s$ for zooming in. As can be seen in the first three column in \cref{fig:geo_transform}, CREPS can perform those simple transformations with consistent details. Notably, CREPS can extrapolate the points outside the original image boundary, although it has never been trained on such input coordinates.

It is tricky for CREPS to handle complex transformations such as rotation or distortion since CREPS only takes in a row and a column coordinate input. Instead of producing the target image in one run, we can execute CREPS multiple times to generate different parts of the output image, then combine them. The simplest algorithm is to sample each target pixel per run by setting a single value for $e^r$ and $e^c$. However, that algorithm is too slow, which requires 262k runs to produce a single $512\times512$ image. A faster way is to sample the target image row-by-row. Assuming we need to generate an image $I$ with the input latent $z$ and the target pixels' normalized coordinates $\{(r_{ij}, c_{ij})\}_{i=\overline{1,H}, j=\overline{1,W}}$. We can produce each row $I_i$ of the target image by generating an intermediate image I' using the input coordinates $e^r = [r_{ij}]_{j=\overline{1,W}}$ and $e^c = [c_{ij}]_{j=\overline{1,W}}$ and sample its diagonal $I_i = diag(I')$. We provide two examples with rotation and elastic distortion in the last two columns of \cref{fig:geo_transform}. Both images are correctly transformed with unchanged content. Additional qualitative result on geometric transformation can be viewed at \href{https://drive.google.com/file/d/1PjMX90L_p74OBUkvADm57s7LMUh-iVgR/view}{here}

\subsection{Feature analysis}\label{sec:feat_vis}
\begin{figure}[t]
    \small
    \begin{center}
        \includegraphics[width=.477\textwidth]{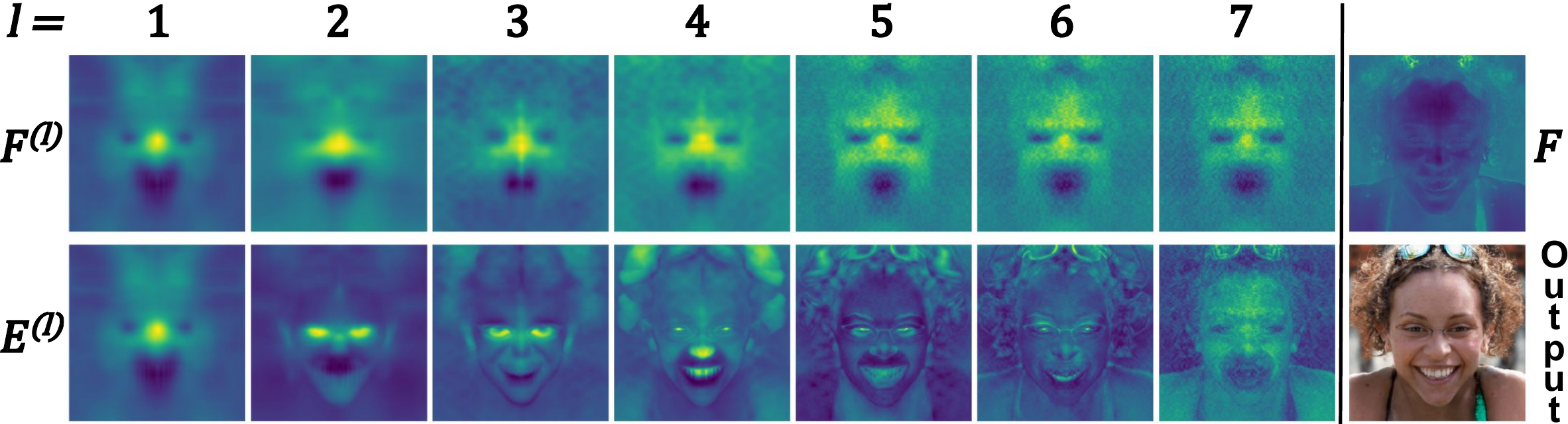}
    \end{center}
    \vskip -0.22in
    \caption{Visualization of the feature maps extracted from our FFHQ-512 model. Each feature map is averaged over all channels.}
    \label{fig:feat}
    \vspace{-2mm}
\end{figure}

We visualize the key feature maps inside our FFHQ-512 model when generating a facial image and provide them in \cref{fig:feat}. The maps include the layer-wise composed features $\{F^{(l)}\}_{l=\overline{1,N}}$, the corresponding layer-wise fused maps $\{E^{(l)}\}_{l=\overline{1,N}}$, and the final feature map $F$ (see Eq. \ref{eq:fuse_1}-\ref{eq:fuse_3}). Thanks to the asymmetric fusion scheme, the model seems to synthesize the output in a coarse-to-fine manner. The early composed feature maps are smooth and focus on the global structure, while the later ones focus on sharp details. Although each composed feature map $F^{(l)}$ is quite simple, the network can represent complex content by fusion.

\subsection{Limitation}\label{sec:limitation}
\begin{figure}[t]
    \small
    \begin{center}
        \includegraphics[width=.47\textwidth]{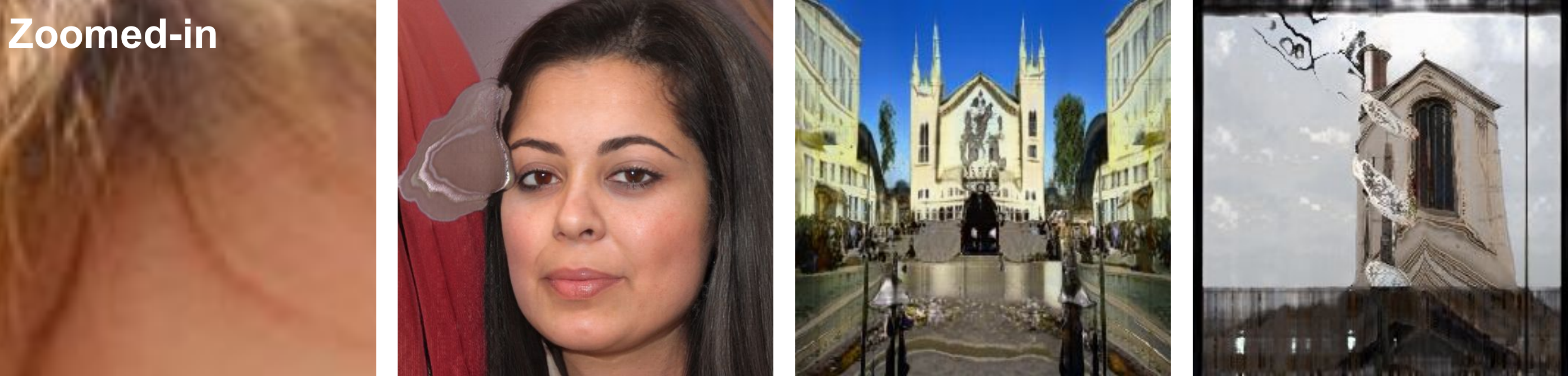}
    \end{center}
    \vskip -0.22in
    \caption{Samples of the most common kinds of artifacts on different datasets. They are best described as repeating/wavy patterns, vertical symmetry, and glowing blobs. Left-most image is cropped and zoomed-in from a full-face image.}
    \label{fig:artifact}
    \vspace{-5mm}
\end{figure}

Being a fully-connected generator, CREPS shares the same limitation with other similar work, which is the lack of spatial bias since each pixel is independently generated. Hence, some spatial-related artifacts occasionally occur in our generated images (\cref{fig:artifact}). A potential cause is the sine activation at the beginning, producing repeating patterns and vertical symmetry of the output. We also note that some samples contain a noticeable blob that is completely out-of-domain. We found CIPS facing the same problem, and the root cause can be the missing spatial guidance from neighboring pixels and the effect of Leaky-RELU activations which strengthens the isolation of some pixel regions. 

\section{Conclusion}\label{sec:conclusion}
In this paper, we present a new architecture named \textbf{CREPS}, a cost-effective and scale-equivariant generator that can synthesize images with any target resolution. Our key contributions are an INR-based design, a thick bi-line representation, and a layer-wise feature composition scheme. While being more memory-efficient, our CREPS models can produce highly realistic images and surpass the INR-based model CIPS in most cases. CREPS also offers the best scale consistency by keeping image details unchanged when varying the output resolution. We conducted several experiments to explore some attractive properties of this fully-connected generator and discussed CREPS's applications in various scenarios. Future development of our approach can be eliminating artifacts mentioned in \cref{sec:limitation} and further improving the quality of our samples.

{\small
\bibliographystyle{ieee_fullname}
\bibliography{egbib}
}

\clearpage
\appendix

\section{Implementation details}\label{sec:sup_implementation}

\subsection{Architecture details}
\label{sec:sup_architecture}

In this section, we describe in detail the implementation of each component in our proposed method.

\minisection{Synthesis block} As reported in the main paper, this block is largely identical to the blocks in \cite{Tero20} with some minor modifications. First, we change the kernel size of the convolution operator from $3 \times 3$ to $1 \times 1$ one. Second, we dismiss the use of small injection noise to the feature output as it is against our objective of scale-invariant generation. Third, we double the number of channels in this block compared to \cite{Tero20} to improve the capacity of the bi-line features. This block both receives and outputs bi-line features.

\minisection{Refinements block} consists of two synthesis blocks with the hidden width of 128 and 64, respectively (\cref{tab:refinement_block}). Instead of bi-line features, this block input and output are both 2D features; the input feature map is decoded from previous bi-line features. The residual output of each synthesis block will be an RGB image in the shape of $3 \times R \times R$.

\begin{table}[b]
\centering
\begin{tabular}{lll}
\hline
Layer                   & Input Shape & Output Shape   \\ \hline
SynthesisBlock(32, 128) & $32 \times R \times R$  & $128 \times R \times R$   \\
ToRGB(128, 3)           & $128 \times R \times R$ & $3 \times R \times R$     \\
SynthesisBlock(128, 64) & $32 \times R \times R$  & $64 \times R \times R$     \\ 
ToRGB(64, 3)            & $64 \times R \times R$  & $3 \times R \times R$      \\ \hline
\end{tabular}
\centering
\caption{Structure of Refinements Block. \label{tab:refinement_block}}
\end{table}

\minisection{Decoder block} is a stack of fully-connected layers with LeakyReLU activations in-between. The structure of this block is illustrated in \cref{tab:decoder_block}.

\begin{table}[b]
\centering
\begin{tabular}{lll}
\hline
Layer           & Input Shape & Output Shape \\ \hline
Fusion          & $32 \times R \times 2D$ & $R \times R \times 32$   \\
Linear(32, 64)  & $R \times R \times 32$  & $R \times R \times 64$   \\
Linear(64, 128) & $R \times R \times 64$  & $R \times R \times 128$  \\
Linear(128, 64) & $R \times R \times 128$ & $R \times R \times 64$   \\
Linear(64, 32)  & $R \times R \times 64$  & $R \times R \times 32$   \\
Permute         & $R \times R \times 32$  & $32 \times R \times R$   \\ \hline
\end{tabular}
\centering
\caption{Structure of Decoder Block. \label{tab:decoder_block}}
\end{table}

\subsection{Transfer learning details}
\label{sec:sup_transfer_learning}

Similar to Karras et al. \cite{Tero20b}, we train MetFaces and AFHQ-Dog (next section) with adaptive discriminator augmentation (ADA) \cite{Tero20b} using weights trained on FFHQ-512. Even though our FFHQ was trained with resolution $512\times512$ only, we can easily train on resolution $1024\times1024$ simply by doubling the length of row and column coordinates $e^r$ and $e^c$. The transfer learning results are reported in \cref{sec:sup_quantitative}.

\subsection{Training config}
\label{sec:sup_training_config}

To train our models, we start with the batch size of 128 and gamma of 0.5 for resolution $128 \times 128$. For higher resolution, we decrease the batch size and increase gamma to further stablilize the training. Specifically, for resolution $512 \times 512$, we use 32 and 10 for batch size and gamma respectively. Lastly, we  set the batch size and gamma as 8 and 32 for resolution $1024 \times 1024$

\section{Additional Quantitative Results}\label{sec:sup_quantitative}

\subsection{Transfer learning results on AFHQ-Dog}

Besides MetFaces, we conduct a further experiment to verify the adaptability of our model from FFHQ to AFHQ-Dog. AFHQ-Dog consists of 4677 facial images of various dog breeds at resolution $1024\times1024$. Following prior works \cite{Tero20b}, we directly use the weight of CREPS trained on FFHQ and continue the training on AFHQ-Dog. Our model achieved an FID score of 9.7, which is slightly higher than the FID score of StyleGAN2-ADA (7.4). However, qualitatively, the images generated by this model are of good quality as illustrated in \cref{fig:sup_afhq}.

\subsection{Comparison With AnyresGAN and ScaleParty}
We provide an additional comparison in terms of FID score with two prior works that support any-scale image synthesis, including AnyresGAN \cite{chai2022anyresolution} and ScaleParty \cite{Evangelos22}, in \cref{tab:sup_quantitative}. Note that both of them make use of spatial convolution, so they are not scale-consistent. Here, the FID scores of AnyresGAN are taken directly from the paper, while those for ScaleParty are re-computed using their publicly available code and pre-trained model.

\begin{table*}[t]
\centering
\begin{tabular}{|c|c|c|c|}
\hline
Generator        & FFHQ-512    &     FFHQ-1024     & LSUN Church-256  \\ \hline
ScaleParty       & 6.23$^\dag$ &     10.91$^\dag$  &    N/A           \\ \hline
AnyresGAN        & 3.71$^*$    &     4.06$^*$      &    3.84$^*$      \\ \hline
CREPS (ours)     & 4.43        &     4.09$^\ddag$  &    5.50          \\ \hline
\end{tabular}
\centering
\vspace{-2mm}
\caption{Comparison of our method against other works in FID metric. `$^*$' means the result is taken from original paper \cite{chai2022anyresolution}. `$^\dag$' means the result is obtained by re-computing the score using the code from author. `$^\ddag$' means the result is obtained by scaling the output resolution of the FFHQ-512 model. N/A means the pretrained weight for this dataset is not released.
\label{tab:sup_quantitative}}
\vspace{-2mm}
\end{table*}

\section{Additional qualitative results}\label{sec:sup_qualitatve}

\subsection{Super-resolution comparison}
\label{sec:sup_super_resolution}

\begin{figure*}[t]
    \small
    \begin{center}
        \includegraphics[width=.85\textwidth]{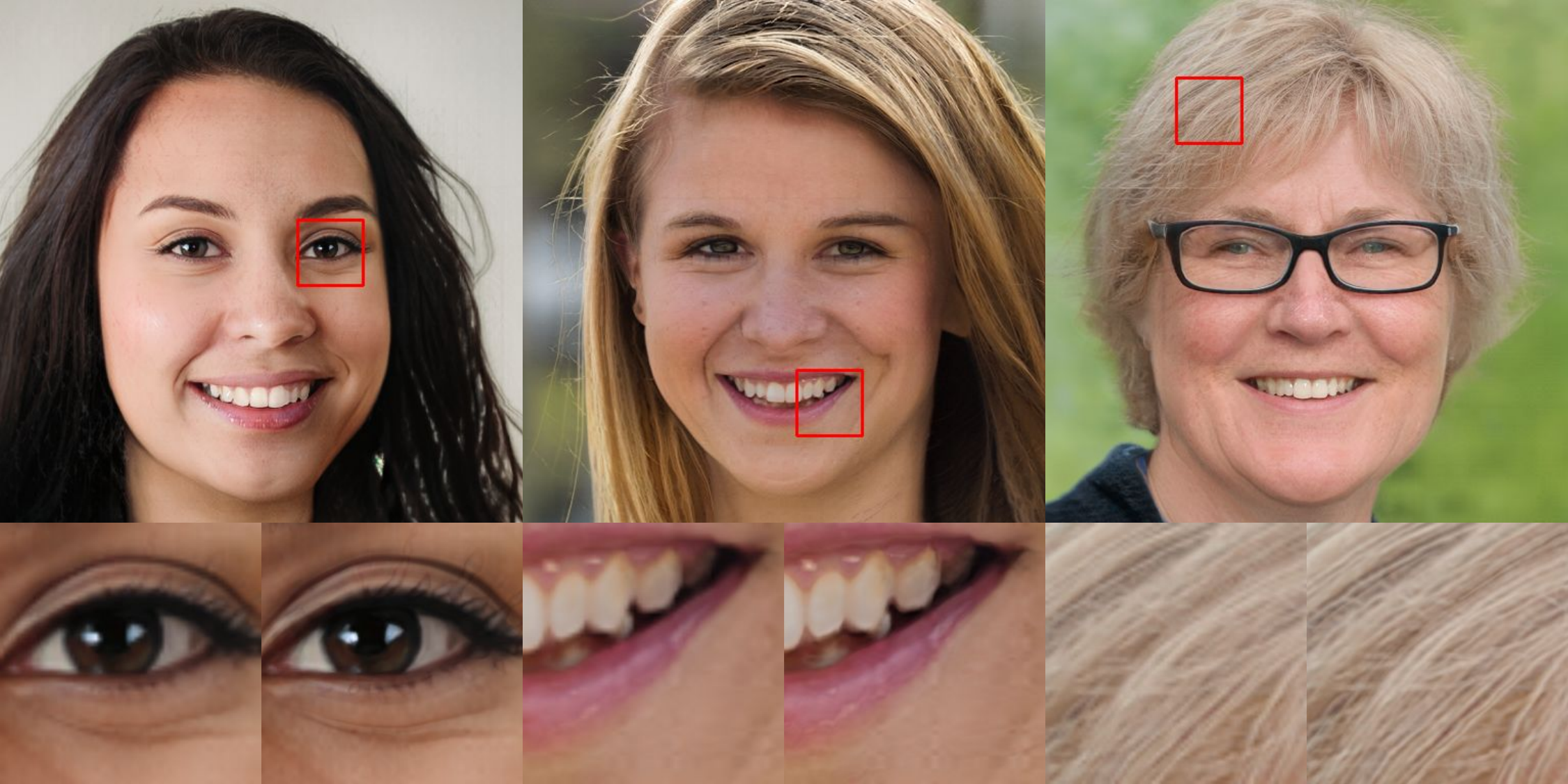}
    \end{center}
    \vskip -0.22in
    \caption{Comparison of CREPS high-resolution image synthesis with Lanzcos upsampling on FFHQ. Top: Images synthesized by CREPS at resolution $512\times512$. Bottom left: the crop at resolution $512\times512$, upscaled with Lanczos upsampling. Bottom right: the corresponded crop of CREPS at resolution $2048\times2048$.}
    \label{fig:sup_highres}
    \vskip -0.15in
\end{figure*}

By scaling the length of row and column coordinates $e^r$ and $e^c$, CREPS can not only generate higher output resolution but also produce finer details. As shown in \cref{fig:sup_highres}, the crop of an image generated by scaling the coordinate of CREPS from 512 to 2048 has more details than directly applying Lanczos upsampling on the corresponding image generated at resolution $512\times512$.

\subsection{Additional image generation results}
\begin{figure*}[t]
    \small
    \begin{center}
        \includegraphics[width=.85\textwidth]{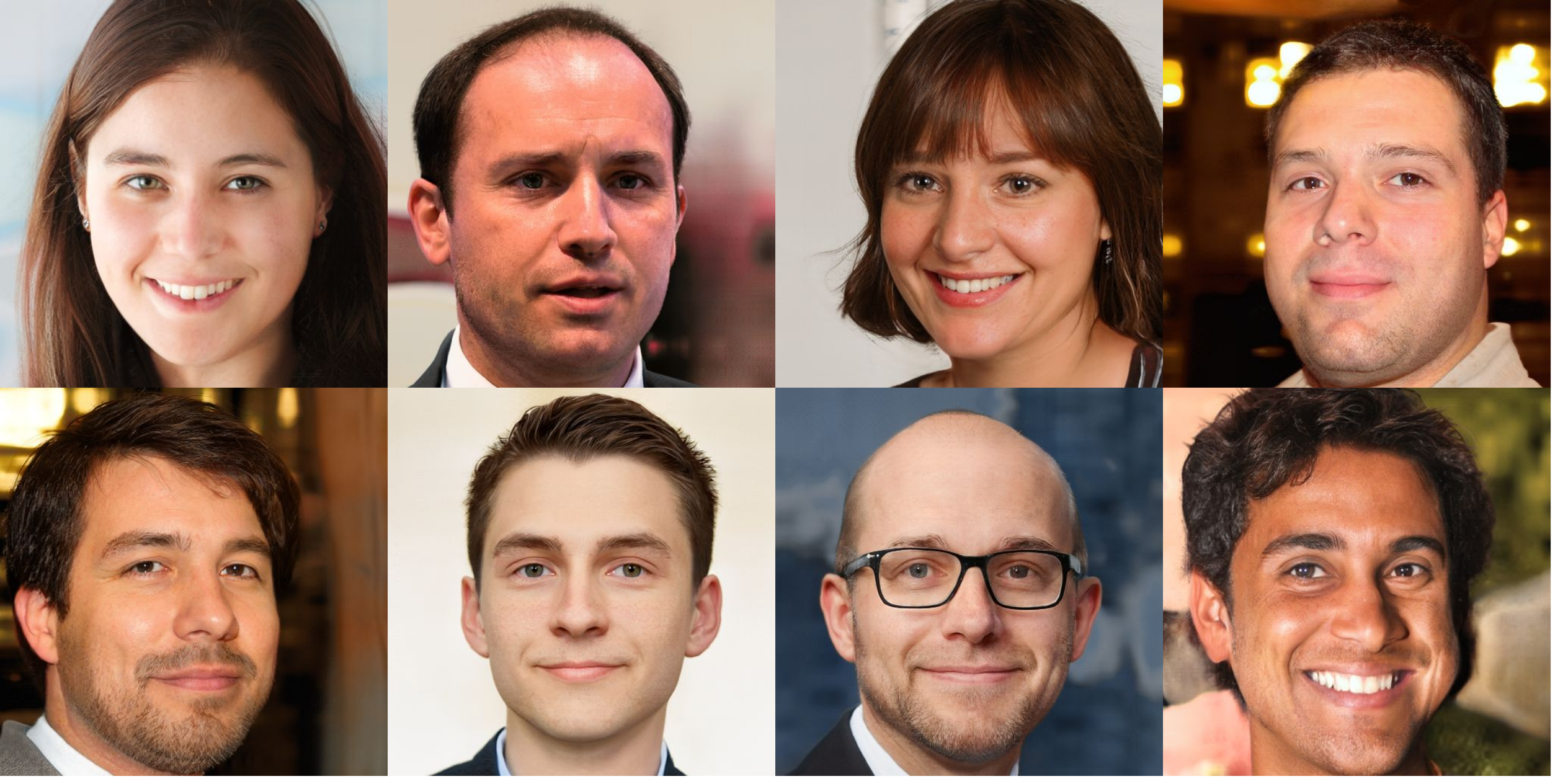}
    \end{center}
    \vskip -0.22in
    \caption{Sample images generated by our models on FFHQ resolution $512\times512$}
    \label{fig:sup_ffhq}
    \vskip -0.15in
\end{figure*}

\begin{figure*}[t]
    \small
    \begin{center}
        \includegraphics[width=.85\textwidth]{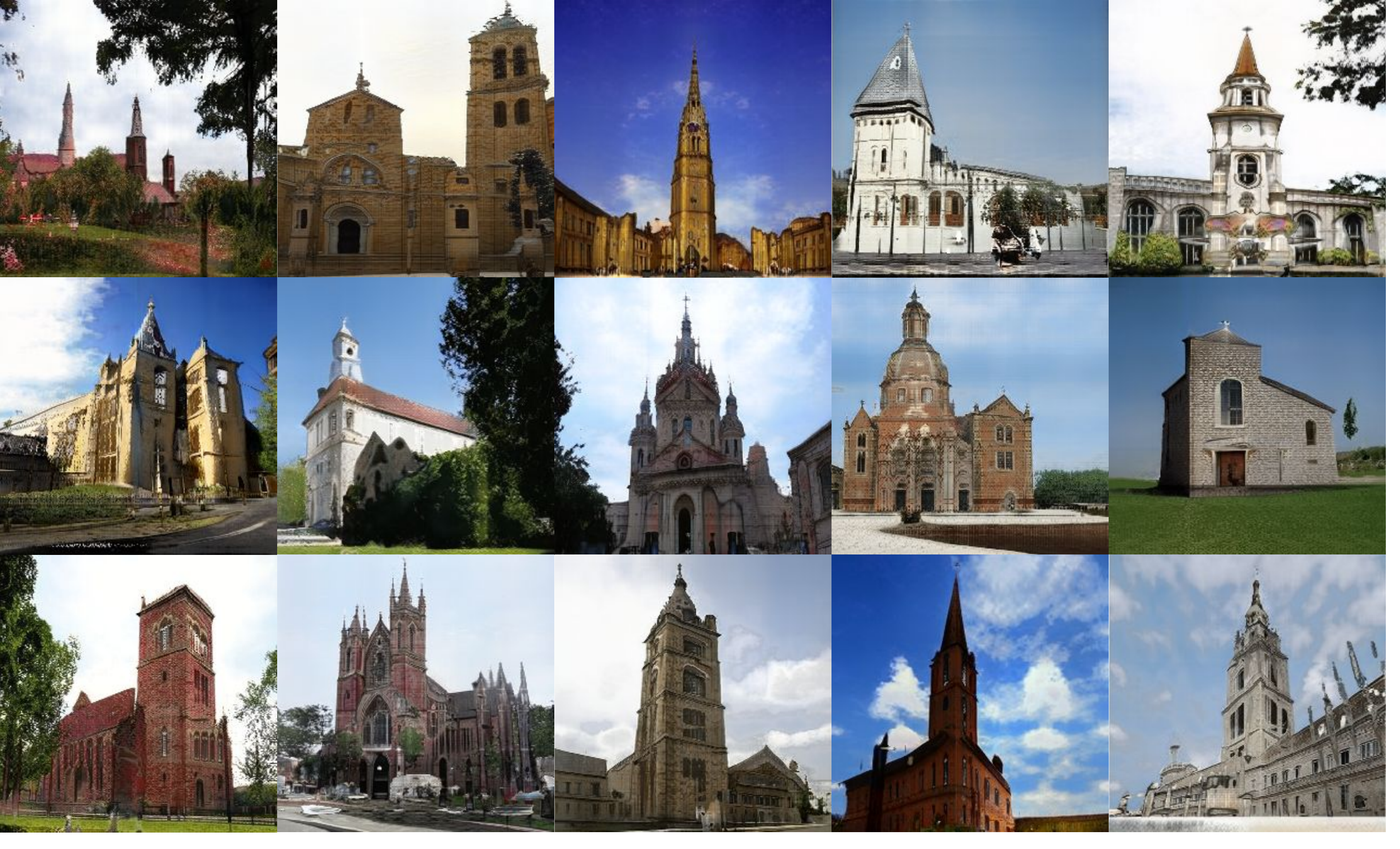}
    \end{center}
    \vskip -0.22in
    \caption{Sample images generated by our models on LSUN Church resolution $256\times256$}
    \label{fig:sup_lsun}
    \vskip -0.15in
\end{figure*}

\begin{figure*}[t]
    \small
    \begin{center}
        \includegraphics[width=.85\textwidth]{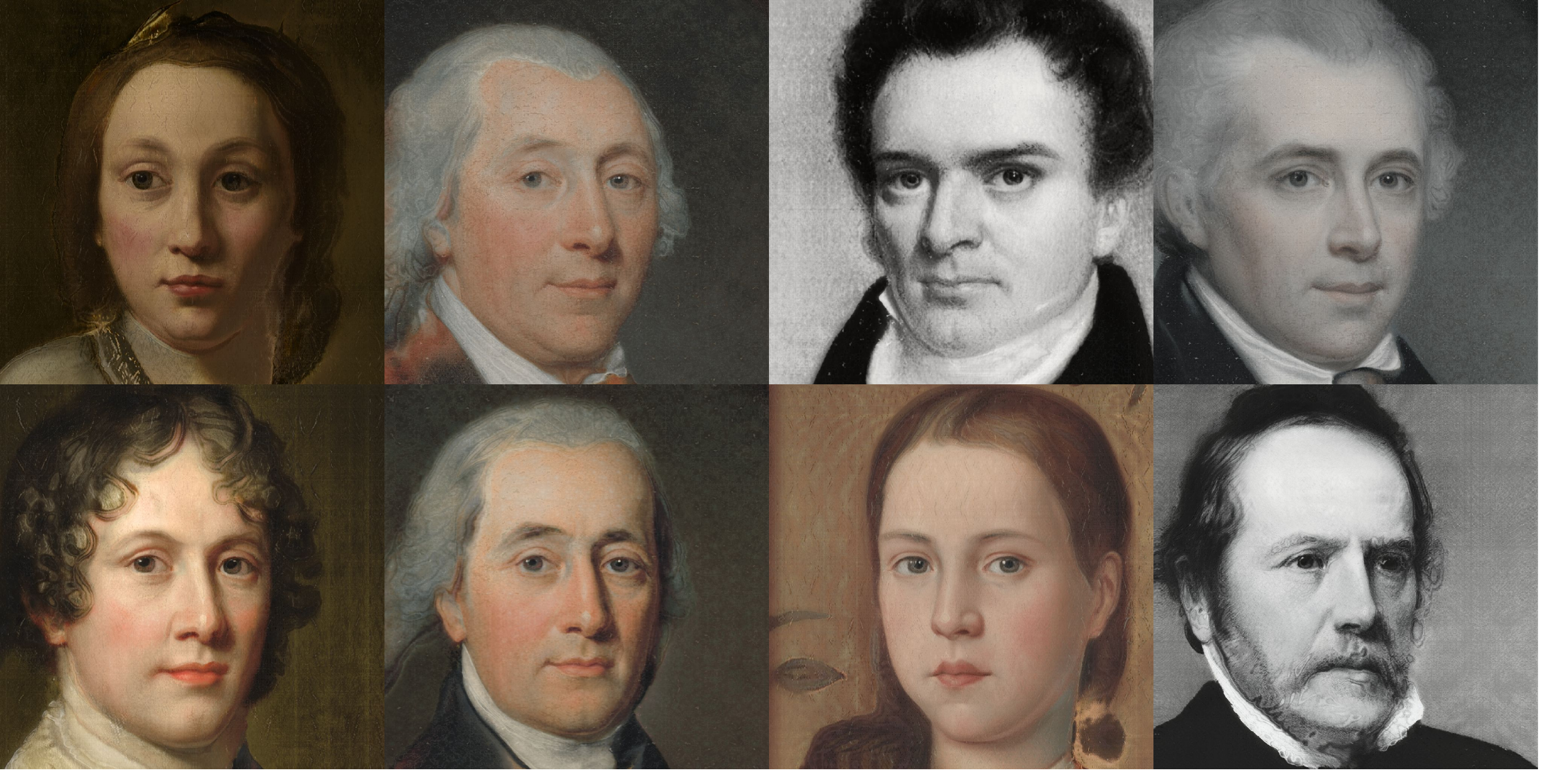}
    \end{center}
    \vskip -0.22in
    \caption{Sample images generated by our models on MetFaces resolution $1024\times1024$}
    \label{fig:sup_metface}
    \vskip -0.15in
\end{figure*}

\begin{figure*}[t]
    \small
    \begin{center}
        \includegraphics[width=.85\textwidth]{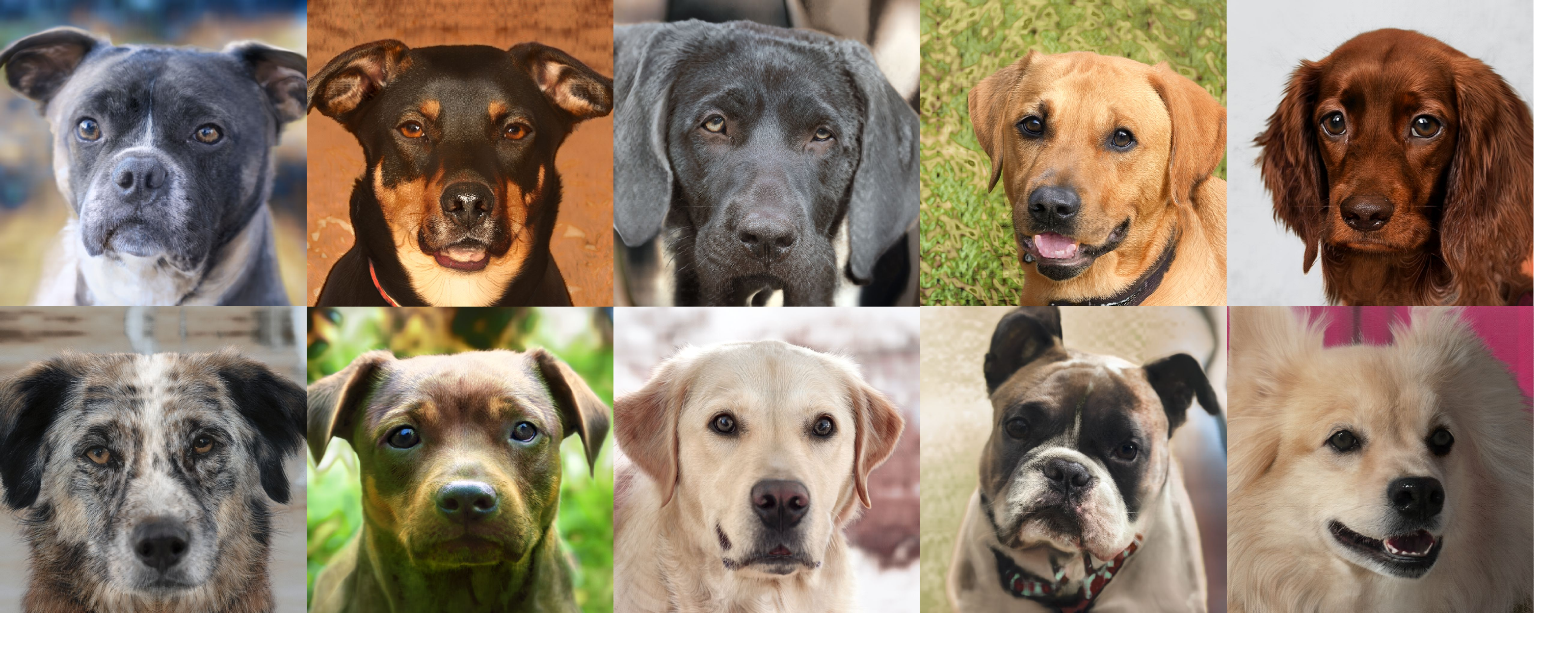}
    \end{center}
    \vskip -0.22in
    \caption{Sample images generated by our models on AFHQ-Dog resolution $1024\times1024$}
    \label{fig:sup_afhq}
    \vskip -0.15in
\end{figure*}

We provide additional results generated by CREPS on FFHQ and LSUN-Church in \cref{fig:sup_ffhq,fig:sup_lsun}. We further verify that our proposed bi-line representation does not limit the capacity of our models by performing transfer learning from FFHQ-512 to MetFaces and AFHQ-Dog. The results are shown in \cref{fig:sup_metface} and \cref{fig:sup_afhq}, respectively.

\end{document}